# An Evasion Attack against Stacked Capsule Autoencoder

Jiazhu Dai, Siwei Xiong

*School of Computer Engineering and Science, Shanghai University, Shanghai 200444, China*

**Abstract:** Capsule networks are a type of neural network that use the spatial relationship between features to classify images. By capturing the poses and relative positions between features, this network is better able to recognize affine transformation and surpass traditional convolutional neural networks (CNNs) when handling translation, rotation and scaling. The stacked capsule autoencoder (SCAE) is a state-of-the-art capsule network that encodes an image in capsules, which each contain poses of features and their correlations. The encoded contents are then input into the downstream classifier to predict the image categories. Existing research has mainly focused on the security of capsule networks with dynamic routing or EM routing, while little attention has been given to the security and robustness of SCAEs. In this paper, we propose an evasion attack against SCAEs. After a perturbation is generated based on the output of the object capsules in the model, it is added to an image to reduce the contribution of the object capsules related to the original category of the image so that the perturbed image will be misclassified. We evaluate the attack using an image classification experiment, and the experimental results indicate that the attack can achieve high success rates and stealthiness. This finding confirms that the SCAE has a security vulnerability that allows for the generation of adversarial samples without changing the original structure of the image to fool the classifiers. Our work seeks to highlight the threat of this attack and focus attention on SCAE security.

## 1. Introduction

Image recognition is a popular research topic in machine learning, and convolutional neural networks (CNNs) are among the major methods addressing this task. CNNs abstract images into local features through operations, such as convolution and pooling, and the local features are then used as evidence for identification. CNNs are able to meet the requirements of common image recognition tasks, but they cannot handle images after affine transformation, such as rotation; thus, CNNs are easily affected by adversarial attacks [1].

A capsule network is a type of neural network that is designed to improve the performance of traditional CNNs when handling images after affine transformation. Based on abstraction, the capsule network further analyzes the spatial relationship between features to promote the reliability of the classification. Capsule networks have been developed through three main stages: the dynamic routing capsule network proposed by Sabour et al. [2] in 2017, the EM routing capsule network proposed by Hinton et al. [3] in 2018, and the stacked capsule autoencoder (SCAE) proposed by Kosiorek et al. [4] in 2019.

The SCAE is a state-of-the-art capsule network that uses autoencoders instead of a routing structure. First, the poses of features and the relationship between features are extracted from the image. Then, they are combined and encoded into capsules. Finally, the predicted result of the image is obtained by inputting the output of the capsules into the classifier. One of the highlights of the SCAE is the unsupervised classification on the capsule network, which uses bipartite graph matching [5] to find the permutation of cluster indices after finding multiple clusters using the k-means approach.

Recent studies have found that capsule networks also face security threats [6–9]. However, these studies only focus on capsule networks based on dynamic routing. To the best of our knowledge, the potential security vulnerabilities of SCAE have not been studied. In this paper, we address this issue. The contributions of the paper can be summarized as follows:

1. We propose an evasion attack against the SCAE in which the attacker can compute the perturbation based on the output of the capsules in the model to generate adversarial samples that lead to misclassification of the SCAE;

2. The attack achieves a high attack success rates on various datasets, which confirms that the SCAE has a security vulnerability that allows for the generation of adversarial samples without changing the original structure of the image to fool the unsupervised classifier in the SCAE.

The remainder of this paper is arranged as follows: Section II introduces the related works, Section III describes the architecture of the SCAE and explains its operations, Section IV presents our attack method and algorithm in detail, Section V describes the experiments and the results, and Section VI provides a summary and briefly presents our future work.

## 2. Related Works

### 2.1 Capsule Network

Capsule networks have been developed through three main stages: the dynamic routing capsule network proposed by Sabour et al. [2] in 2017, the EM routing capsule network proposed by Hinton et al. [3] in 2018, and the stacked capsule autoencoder (SCAE) proposed by Kosiorek et al. [4] in 2019.

The SCAE, as a state-of-the-art capsule network, uses autoencoders instead of a routing algorithm and conducts both supervised and unsupervised classification at the same time. The whole model is composed of a part capsule autoencoder (PCAE) and an object capsule autoencoder (OCAE). After the image is input into the model, the parts of each object in the image are extracted by the PCAE and then combined into whole objects by the OCAE. The classifier makes predictions according to the presence of different parts and objects. Each capsule in the PCAE contains a six-dimensional pose vector, a one-dimensional presence probability, and an n-dimensional attribute vector. The OCAE uses a set transformer [10] to encode part capsules into object capsules. Regarding the classifiers, the SCAE uses linear classifiers for supervised classification and k-means classifiers for unsupervised classification.

The main contribution of the SCAE is that it provides a new learning method using the PCAE to segment images into multiple parts and the OCAE to make the parts into whole objects. This method considers the spatial relationship between features and the variety of representations of similar features; therefore, it is less vulnerable to random perturbation. The SCAE can achieve a high unsupervised classification accuracy of 98.7% on the MNIST dataset.

### 2.2 Poisoning Attacks and Evasion Attacks

The security threats in machine learning can be categorized into two types: poisoning attacks and evasion attacks.

Poisoning attacks occur during training when an attacker adds elaborately constructed malicious samples to the training set to manipulate the behavior of the model at test time, thus causing the model to output the attacker's expected results for specific samples or reducing the classification accuracy of the model [11–20].

Evasion attacks occur during the test phase when an attacker adds a carefully constructed perturbation to the clean sample to form a malicious sample. Its appearance is generally consistent with that of the clean sample, although the model will misclassify it or make a prediction specified by the attacker [21–33]. The adversarial attack proposed in this paper is an evasion attack.

### 2.3 Security Threats of the Capsule Network

After the emergence of the capsule network, research on its security focused on the dynamic routing capsule network. Jaesik [6] provided a variety of successful methods of adversarial attacks on capsule networks. Michels et al. [7] proved that the ability of encapsulated networks to resist white-box attacks is not better than that of traditional CNNs. Marchisio et al. [8] designed a black-box attack algorithm against a capsule network and verified its effectiveness on the German Traffic Sign Recognition Benchmark (GTSRB) dataset. De Marco [9] proved that capsule networks with different scales are vulnerable to adversarial attacks to varying degrees. The SCAE has a different structure than previous capsule networks, and its robustness also needs to be analyzed. To the best of our knowledge, few research reports have focused on SCAE. In this paper, we study the vulnerability of the SCAE to adversarial attacks to help improve its robustness to such security threats.

## 3. Stacked Capsule Autoencoder

The complete structure of the SCAE is shown in Figure 1. The SCAE treats an image as a composition of multiple objects that each consist of multiple parts. These parts and objects will be analyzed by two main units of the model, that is, the part capsule autoencoder (PCAE) and the object capsule autoencoder (OCAE). First, after the image is input into the model, a CNN will be used by the PCAE to extract the poses, presences and features of parts that constitute the objects in the image, which are stored in part capsules. Each part capsule represents a part that may exist in the image. Next, the OCAE uses a set transformer [10] to conduct the autoencoding operation, which combines these scattered parts into complete objects, calculates the pose and presence of each object, and stores them in object capsules. Each object capsule represents an object that may exist in the image. Finally, the classifier will use the capsule output encoded by the OCAE to predict the label of the image.

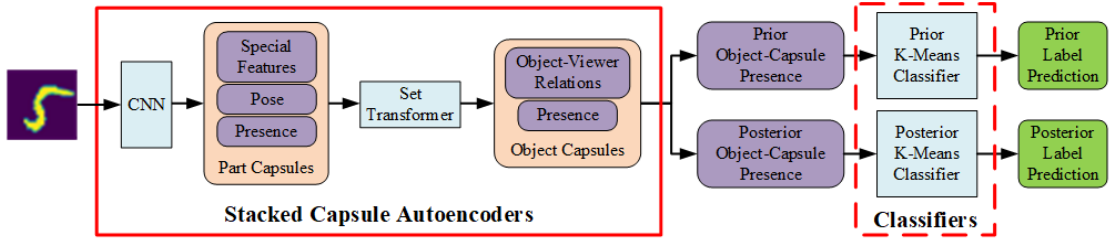

Figure 1 SCAE architecture.

There are two forms of SCAE output: the prior object-capsule presence with dimension $[B, K]$ and the posterior object-capsule presence with dimension $[B, K, M]$, where $B$ is the batch size, $K$ is the number of object capsules, and $M$ is the number of part capsules. The output of the SCAE is the probability of existence of each object in the image, and its value ranges from 0 to 1. The classifier can select any one of the above outputs for k-means clustering and then use bipartite graph matching [5] to find the mapping permutation between clustering labels and ground truth labels. During the test stage, the output of the SCAE is input into the classifier to obtain the predicted label of the image. It should be noted that if the posterior object-capsule presence is used, dimension $M$ needs to be reduced and its values need to be summed to ensure shape consistency. Our attack in this paper targets the classification with the prior k-means classifier and posterior k-means classifier.

## 4. Proposed Evasion Attack

The proposed evasion attack is a white-box attack that consists of two steps. In the first step, we identify the object capsule subset $S$ that contributes most to the original label according to the output of the SCAE model $E$ on input image $x$, and obtain the target function $f(x + p)$ for computing the perturbation, which is described in detail in subsection 4.1. In the second step, inspired by [25,26,33], we design the following three algorithms to compute the minimum perturbation $p$, which aim to decrease the sum of the output of the object capsules belonging to subset $S$ so that the perturbed images are misclassified:

- **Computing the Perturbation with Gradient Direction Update (GDU):** In this

algorithm, the adversarial sample is iteratively updated according to the direction of gradient of the target function $f(x + p)$;

- **Computing the Perturbation with Pixel Saliency to Capsules (PSC)**: In this algorithm, we iteratively choose and modify a pair of pixels which contributes most to the value of the target function $f(x + p)$ until the adversarial sample cannot be classified correctly;
- **Computing the Perturbation with Optimizer (OPT)**: In this algorithm, an optimizer is used to minimize the value of the target function $f(x + p)$ so as to generate the perturbation $p$ which causes misclassification.

The detail of the three algorithms is presented in subsection 4.2.

## 4.1 Identifying the Object Capsule Subset

For image $x$ and the SCAE model $E$, let $E(x)$ be the $K$-dimensional object-capsule presence that the SCAE outputs on $x$ and let $C(E(x))$ be the final classification result of $x$ on the classifier $C$. The goal of our attack is to find a perturbation $p$ with the minimum $\|p\|_2$ that makes the classifier $C$ misclassify $E(x + p)$. This problem is defined as follows:

$$\begin{aligned}\text{Minimize} \quad & \|p\|_2 \\ \text{s.t.} \quad & C(E(x+p)) \neq C(E(x)) \\ & x + p \in [0,1]^n\end{aligned} \quad (1)$$

Because of the sparsity loss used during training, different capsules are often associated with different classes. For an input image, only object capsules related to the label of the image will be activated and output with high presence, while the irrelevant object capsules remain inactive. This feature makes it possible to find the object capsule subset $S$ related to the image. To misclassify a perturbed image $x + p$, that is, $C(E(x + p)) \neq C(E(x))$, we can lower the output of the object capsule subset $S$ to reduce the contribution of this output to the original classification. In other words, we want the SCAE to mistakenly believe that "the objects that belong to the real category do not appear in the image", which is shown in Formula (2):

$$\text{Minimize} \quad f(x + p) = \sum_{i \in S} E(x + p)_i \quad (2)$$

For a well-trained SCAE model, the output object-capsule presence should be close to 1 or 0; that is, the SCAE can accurately distinguish whether the objects represented by each capsule appear in the image. According to this feature, we compute the $K$-dimensional object-capsule presence $E(x)$ for $x$, calculate the average presence $\overline{E(x)} = \frac{1}{K}\sum_{i=1}^{K} E(x)_i$, and finally obtain the activated object capsule subset $S = \{i | E(x)_i > \overline{E(x)}\}$.

## 4.2 Computing the Perturbation

Based on Formula (2), we design three algorithms to compute the minimum perturbation $p$ which aims to decrease the sum of the output of the object capsules belonging to subset $S$ and cause the perturbed image to be misclassified.

### 4.2.1 Computing the Perturbation with Gradient Direction Update

In this algorithm, we compute the sign of the gradient of Formula (2) to the image $x$, and obtain the direction to update the value of each pixel, which decreases $\sum_{i \in S} E(x)_i$. The adversarial sample is generated in this direction:

$$x_0^{adv} = x, \qquad x_{N+1}^{adv} = \text{clip}_{0,1}\left(x_N^{adv} - \alpha \cdot \text{sign}\left(\nabla_{x_N^{adv}} f(x_N^{adv})\right)\right) \tag{3}$$

where $x_N^{adv}$ is the generated adversarial sample based on image $x$, $\alpha > 0$ is a hyperparameter that represents the step of change in each iteration, and $\nabla_x f(x)$ represents the gradient of $x$ to $f(x)$. The pixel values of the generated sample are clipped between 0 and 1. The whole procedure is shown in Algorithm 1.

---

**Algorithm 1:** Generating the Perturbation with Gradient Direction Update

1: **Input:** Image $x$, SCAE model $E$, classifier $C$, hyperparameter $\alpha$, the number of iterations $n_{iter}$.
2: **Output:** Perturbation $p$.
3:
4: Initialize $x^{adv} \leftarrow x$, $x_0^{adv} \leftarrow x$, $\mathcal{L}_p \leftarrow +\infty$.
5: $S \leftarrow \left\{ i \,\middle|\, E(x)_i > \frac{1}{K} \sum_{i=1}^{K} E(x)_i \right\}$
6: **for** $i$ in $n_{iter}$ **do**
7:     /* Compute next $x_i^{adv}$ */
8:     $x_{i+1}^{adv} \leftarrow \text{clip}_{0,1}\left(x_i^{adv} - \alpha \cdot \text{sign}\left(\nabla_{x_i^{adv}} \sum_{j \in S} E(x_i^{adv})_j\right)\right)$
9:     /* Judge whether the current result is the best one */
10:     **if** $C\left(E(x_{i+1}^{adv})\right) \neq C(E(x))$ **and** $\|x_{i+1}^{adv} - x\|_2 < \mathcal{L}_p$ **do**
11:         $x^{adv} \leftarrow x_{i+1}^{adv}$
12:         $\mathcal{L}_p \leftarrow \|x_{i+1}^{adv} - x\|_2$
13:     **end if**
14: **end for**
15: $p \leftarrow x^{adv} - x$
16: **return** $p$

---

### 4.2.2 Computing the Perturbation with Pixel Saliency to Capsules

In this algorithm, we iteratively choose a pair of pixels that contributes most to the value of Formula (2). The amount of contribution of a pixel pair $(p, q)$ is named pixel saliency, which is

defined as follows:

$$\delta_1 = \sum_{i=p,q} \left( \nabla_{x_i} \sum_{j \in S} E(x)_j \right)$$
$$\delta_2 = \sum_{i=p,q} \left( \nabla_{x_i} \sum_{j \notin S} E(x)_j \right) \quad (4)$$

where $\delta_1$ represents the influence on the sum of the output of the object capsule subset $S$ when the pixel pair $(p, q)$ changes, and $\delta_2$ represents the influence on the sum of the output of the object capsules not belonging to $S$. The pixels for modification are picked according to the following rule:

$$(p^*, q^*) = \arg\max_{(p,q)} \left( -\delta_1 \times \delta_2 \times (\delta_1 > 0) \times (\delta_2 < 0) \right) \quad (5)$$

where $\delta_1 > 0$ means the sum of the output of the capsules in subset $S$ increases with the values of the pixels, and $\delta_2 < 0$ means the sum of the output of the other capsules decreases when the values of the pixels increase. Therefore, by reducing the value of the pixels, we can decrease the sum of the output of the capsules in subset $S$.

Before the iteration, a pixel search domain $\Gamma$ is initialized including all pixels whose values are above 0, which are available for modification. We use an iterative algorithm to compute the adversarial sample. In each iteration, we obtain the pair of pixels with the largest pixel saliency from $\Gamma$, and reduce the values of the two pixels by $\alpha$, which is a positive hyperparameter that represents the step of change. During the iteration, the modification of $\Gamma$ follows this principle: if the value of a pixel reaches 0 after being chosen and decreased, then this pixel will be removed from $\Gamma$ because 0 is the minimum value of a pixel and further subtraction is meaningless. All pixels not in $\Gamma$ are excluded from the algorithm.

The algorithm halts as soon as the current perturbation causes misclassification to avoid affecting more pixels. On the other hand, the algorithm will also stop when $\Gamma$ is empty, which means that the attack fails. The whole procedure is shown in Algorithm 2.

| | |
|---|---|
| **Algorithm 2:** Generating the Perturbation with Pixel Saliency to Capsules | |
| 1: | **Input:** Image $x$, SCAE model $E$, classifier $C$, hyperparameter $\alpha$, the number of iterations $n_{iter}$. |
| 2: | **Output:** Perturbation $p$. |
| 3: | |
| 4: | /* *Initial feature search domain* */ |
| 5: | Initialize $x^{adv} \leftarrow x$, $\Gamma \leftarrow \{p \mid x_p > 0\}$. |

```
 6:    S ← {i | E(x)_i > (1/K) Σ_{i=1}^{K} E(x)_i}
 7:    for i in n_iter do
 8:        /* Choose the two pixels for modification */
 9:        m ← 0
10:        for each pair (p, q) in Γ do
11:            δ_1 = Σ_{i=p,q}(∇_{x_i} Σ_{j∈S} E(x)_j)
12:            δ_2 = Σ_{i=p,q}(∇_{x_i} Σ_{j∉S} E(x)_j)
13:            if −δ_1 × δ_2 × (δ_1 > 0) × (δ_2 < 0) > m do
14:                p_1, p_2 ← p, q
15:                m ← −δ_1 × δ_2
16:            end if
17:        end for
18:        /* Decrease the values of the two pixels */
19:        x_{p_1}^{adv} ← max(x_{p_1}^{adv} − α, 0)
20:        x_{p_2}^{adv} ← max(x_{p_2}^{adv} − α, 0)
21:        if x_{p_1}^{adv} == 0 do
22:            Remove p_1 from Γ.
23:        end if
24:        if x_{p_2}^{adv} == 0 do
25:            Remove p_2 from Γ.
26:        end if
27:        if C(E(x^{adv})) ≠ C(E(x)) or Γ is empty do
28:            break
29:        end if
30:    end for
31:    p ← x^{adv} − x
32:    return p
```

### 4.2.3 Computing the Perturbation with Optimizer

In this algorithm, an optimizer is used to minimize the value of Formula (2). To misclassify a perturbed image $x + p$, that is, $C(E(x + p)) \neq C(E(x))$, we can lower the output of the object capsule subset $S$ by decreasing the value of $f(x + p)$ to reduce their contribution to the original classification. Therefore, the optimization problem in Formula (1) can be transformed into the one in Formula (6):

$$\begin{aligned} \text{Minimize} \quad & \|p\|_2 + \alpha \cdot f(x + p) \\ s.t. \quad & x + p \in [0,1]^n \end{aligned} \quad (6)$$

where $\alpha > 0$ is a suitably chosen hyperparameter, which ensures that the two parts in Formula (6) can be optimized simultaneously. The amount of perturbation $\|p\|_2$ is also taken into consideration and minimized by the optimizer, so that the generated adversarial sample can obtain better stealthiness.

For the box constraints in Formula (6), namely, $x + p \in [0,1]^n$, instead of optimizing over the variable $p$ and clipping the pixel values of $x + p$ directly, we introduce a new variable $p'$ according to the "change of variables" method given by Carlini et al. [25] and optimize over it instead. The relationship between the two variables is as follows:

$$\begin{aligned} w &= \text{arctanh}(2x - 1) \\ p &= \frac{1}{2}(\tanh(w + p') + 1) - x \end{aligned} \quad (7)$$

As $x \in [0,1]^n$, $w \in (-\infty, +\infty)^n$.[1] For $p' \in (-\infty, +\infty)^n$, $\tanh(w + p') \in (-1,1)^n$, which leads to $x + p \in (0,1)^n$. The core idea of this method is that the original image $x$ is first transformed into the $arctanh$ space and then mapped back into $[0,1]^n$; therefore, the calculated image $x + p$ is always valid.

The full algorithm for computing perturbation consists of the inner iteration and the outer iteration.

In the inner iteration, we use an optimizer to iteratively conduct the process of Formula (6). The hyperparameter $\alpha$ is fixed from beginning to end during the computation. When an inner iteration is finished, we select the perturbation $p$ with the smallest $\|p\|_2$ from those satisfying $C(E(x + p)) \neq C(E(x))$ as the best inner result.

In the outer iteration, we initialize the optimizer, execute a complete inner iteration, and finally update the hyperparameter $\alpha$ to find the most suitable value of $\alpha$ so that the optimizer can obtain perturbation $p$ with a smaller $L_2$ norm. We perform multiple rounds of outer iterations, and finally select the perturbation $p$ with the smallest $\|p\|_2$ among all best inner results as the global optimal result.

The whole procedure is shown in Algorithm 3:

---

**Algorithm 3:** Generating the Perturbation with Optimization

1: **Input:** Image $x$, SCAE model $E$, classifier $C$, optimizer $opt$, hyperparameter $\alpha$, the number of outer iterations $n_{o\_iter}$, the number of inner iterations $n_{i\_iter}$.
2: **Output:** Perturbation $p$.
3:
4: Initialize $p \leftarrow 0$, $\mathcal{L}_p \leftarrow +\infty$, $\alpha$.
5: $S \leftarrow \{i | E(x)_i > \frac{1}{K}\sum_{i=1}^{K} E(x)_i\}$
6: $w \leftarrow \text{arctanh}(2x - 1)$
7: **for** $i$ in $n_{o\_iter}$ **do**
8:     Init $opt$.
9:     $p'_0 \leftarrow \text{rand}(0,1)$

---

[1] During the experiments, we use $\text{arctanh}((2x - 1) * \epsilon)$ to avoid dividing by zero.

```
10:        for j in n_{i_iter} do
11:            x_j^{adv} ← (1/2)(tanh(w + p'_j) + 1)
12:            L ← ||x_j^{adv} − x||_2 + α · Σ_{i∈S} E(x_j^{adv})_i
13:            /* The optimizer computes next p'_j */
14:            p'_{j+1} ← opt(p'_j|L)
15:            x_{j+1}^{adv} ← (1/2)(tanh(w + p'_{j+1}) + 1)
16:            /* Judge if the current result is the best one */
17:            if C(E(x_{j+1}^{adv})) ≠ C(E(x)) and ||x_{j+1}^{adv} − x||_2 < L_p do
18:                L_p ← ||x_{j+1}^{adv} − x||_2
19:                p ← x_{j+1}^{adv} − x
20:            end if
21:        end for
22:        Update α.
23: end for
24: return p
```

The update algorithm of the hyperparameter $\alpha$ is as follows: we use a binary search to find the optimal value for $\alpha$. First, we specify the upper and lower bounds for $\alpha$ as $\alpha_{ub}$ and $\alpha_{lb}$ respectively, and assign $\alpha$ an initial value between them. Then, in the inner iterations, if the algorithm can obtain any perturbation $p$ that satisfies $C(E(x^{adv})) \neq C(E(x))$, let $\alpha_{ub} \leftarrow \alpha$; otherwise, $\alpha_{lb} \leftarrow \alpha$. Finally, we take $\alpha \leftarrow (\alpha_{ub} + \alpha_{lb})/2$ as the new value for $\alpha$.

## 5. Experimental Evaluation

### 5.1 Experimental Setup

The original paper on SCAE [4] showed that the SCAE can achieve 98.7% accuracy on MNIST, while only achieve 55.33% and 25.01% accuracy on SVHN and CIFAR10, which is not high enough to meet the needs of our experiments. ImageNet, which is similar to CIFAR10, is unsuitable for our experiments as well. Based on the performance of the SCAE on various datasets, we finally choose three datasets for the experiments: MNIST, Fashion-MNIST and German Traffic Sign Recognition Benchmark (GTSRB), in order to make our experimental results more convincing. We train three SCAE models [4,34] using the main parameter set as shown in Table 1.

Table 1 SCAEs' main parameters.

| Dataset | MNIST | Fashion MNIST | GTSRB |
|---|---|---|---|
| Canvas size | 40 | 40 | 40 |
| Num of part capsules | 24 | 24 | 40 |
| Num of object capsules | 24 | 24 | 64 |
| Num of channels | 1 | 1 | 3 |
| Template size | 11 | 11 | 14 |
| Part capsule noise scale | 4.0 | 4.0 | 0.0 |
| Object capsule noise scale | 4.0 | 4.0 | 0.0 |
| Part CNN | 2x(128:2)-2x(128:1) | 2x(128:2)-2x(128:1) | 2x(128:1)-2x(128:2) |
| Set transformer | 3x(1-16)-256 | 3x(1-16)-256 | 3x(2-64)-256 |

Most of the parameters are equivalent to those given in the supplemental document of the original paper on SCAE [4]. For part CNN, 2x(128:2) means two convolutional layers with 128 channels and a stride of two. For the set transformer, 3x(1–16)-256 means three layers, one attention head, 16 hidden units and 256 output units. All three models share the same optimizer shown in Table 2.

Table 2 Settings of the optimizer to train the SCAEs.

| Optimizer parameter | Value |
|---|---|
| Algorithm | RMSProp |
| Learning rate | $3 \times 10^{-5}$ |
| Momentum | 0.9 |
| $\epsilon$ | $1 \times 10^{-6}$ |
| Learning rate decay steps | 10000 |
| Learning rate decay rate | 0.96 |
| Batch size | 100 |

According to [4], the value of $k$ of the k-means classifiers is set as 10, which equals the number of the categories.

In particular, when training the SCAE on GTSRB, the performance is poor when all 43 categories are used; thus, we choose only 10 categories to ensure that the classification accuracy meets the requirement of our experiments. Furthermore, we use min-max normalization (shown Formula (8)) to reduce the luminance difference between images. The chosen categories are shown in Figure 2.

$$x_i' = \frac{x_i - \min(x)}{\max(x) - \min(x)} \tag{8}$$

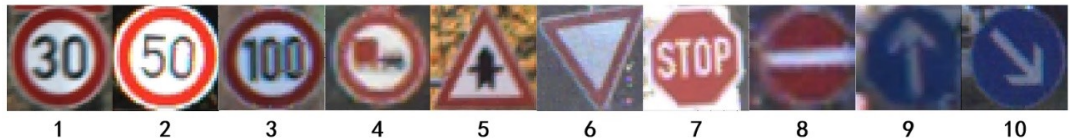

Figure 2 Chosen categories of the German traffic sign recognition benchmark dataset used to train the SCAE model.

The classification accuracy of all three models is given in Table 3:

Table 3 Accuracy of the SCAE models trained on three datasets (in %).

| Dataset | MNIST | Fashion MNIST | GTSRB |
|---|---|---|---|
| Prior k-means classifier | 97.82 | 63.20 | 60.11 |
| Posterior k-means classifier | 97.62 | 63.79 | 56.14 |

## 5.2 Experimental Method

We evaluate our attack on both the prior k-means classifier and the posterior k-means classifier. In each experiment, we choose 5000 samples that can be recognized correctly by the target classifier. The perturbed images are computed using the above three algorithms separately. Then, these images are input into the SCAE, and the encoded results of the SCAE are fed into the corresponding classifier to test the attack success rate (the ratio of the number of samples that are misclassified by the classifier to the total number of all test samples). We compare the three algorithms to find the best algorithm for our evasion attack. The source code of our attack is available at https://github.com/FrostbiteXSW/SCAE_Attack.

The imperceptibility of the perturbation is a challenge on images from the MNIST dataset and the Fashion MNIST dataset because the background of those images is black (the pixel value is zero) and white noise on the black background can be easily noticed. Therefore, we place mask $m$ on the perturbation to confine its scope to the objects and their vicinity rather than the background of the images to improve the stealthiness of the attack. The mask value $m_i$ at each pixel position $i$ is calculated using Formula (9):

$$m_i = \frac{1}{9}\left(x_i + \sum N_8(x_i)\right) \quad (9)$$

where $m_i$ is the average of the corresponding pixel $x_i$ and its eight neighbors $N_8$. For the perturbation value $p_i$ on each pixel $i$, let $p_i' \leftarrow p_i * m_i$, where $m_i$ is used as the weight of variable $p_i$. In the background area $p_i' = 0$, which is because $m_i = 0$; therefore, the background area of the image is not affected by the perturbation.

In the following subsections, we will introduce the setups for each algorithm, and explain how the mask method is applied to them separately.

### 5.2.1 Experimental Settings for GDU

We set the hyperparameter $\alpha$ to 0.05 and the number of iterations to 100. The mask $m_i$ is multiplied by $\alpha \cdot \text{sign}\left(\nabla_{x_N^{adv}} \sum_{i \in S} E(x_N^{adv})_i\right)$ to limit the area affected by the perturbation.

### 5.2.2 Experimental Settings for PSC

We set the hyperparameter $\alpha$ to 0.5 and the number of iterations to 200. The mask $m_i$ is multiplied by the saliency $\delta_1$ and $\delta_2$ to suppress the saliency of particular pixels thus limiting the area affected by the perturbation.

### 5.2.3 Experimental Settings for OPT

We set the initial value of hyperparameter $\alpha$ and its upper and lower bounds to 100, $+\infty$ and 0, respectively. If the upper bound of $\alpha$ is $+\infty$ when updating it, then we simply let $\alpha \leftarrow \alpha \times 10$. The mask $m_i$ is multiplied directly by the generated perturbation. The numbers of inner and outer iterations are limited to 300 and 9, respectively, and the optimizer used to compute the perturbations is shown in Table 4.

Table 4 Settings of the optimizer used to attack the SCAEs.

| Optimizer parameter | Value |
| --- | --- |
| Algorithm | Adam |
| Learning rate | 1.0 |
| $\beta_1$ | 0.9 |
| $\beta_2$ | 0.999 |
| $\epsilon$ | 1x10⁻⁸ |

## 5.3 Results and Discussion

We perform the experiment on each dataset and each classifier with the above three algorithms, and the results are shown in Table 5, Table 6 and Table 7 respectively. The attack success rate is the ratio of the number of samples that are misclassified by the classifier to the total number of all test samples. The average $L_2$ norm of $p$ represents the average amount of the perturbations, and a lower value corresponds to an attack with better stealthiness. The standard deviation of $p$ is used to estimate the stability of the attack. A higher standard deviation means that the performance of the attack fluctuates according to different samples.

Table 5 Experimental results for the MNIST dataset.

| Classifier | Algorithm for computing perturbations | Attack success rate | Average $L_2$ norm of $p$ | Standard deviation of $p$ |
| --- | --- | --- | --- | --- |
| Prior k-means classifier | GDU | 0.9790 | 2.6292 | 1.0797 |
|  | PSC | **1.0000** | 3.7964 | 1.5333 |
|  | OPT | **1.0000** | **1.0839** | **0.6015** |
| Posterior k-means classifier | GDU | 0.8994 | 3.1017 | 1.3118 |
|  | PSC | 0.9994 | 4.4982 | 1.6447 |
|  | OPT | **1.0000** | **1.1838** | **0.6222** |

Table 6 Experimental results for the Fashion MNIST dataset.

| Classifier | Algorithm for computing perturbations | Attack success rate | Average $L_2$ norm of $p$ | Standard deviation of $p$ |
| --- | --- | --- | --- | --- |
| Prior k-means | GDU | 0.9836 | 3.0482 | 1.5615 |
| classifier | PSC | 0.9960 | 4.0945 | 1.8384 |
| | OPT | **1.0000** | **1.4160** | **0.8982** |
| Posterior k-means classifier | GDU | 0.9686 | 3.0768 | 1.6260 |
| | PSC | 0.9622 | 4.0196 | 1.9138 |
| | OPT | **1.0000** | **1.3598** | **0.9856** |

Table 7 Experimental results for the GTSRB dataset.

| Classifier | Algorithm for computing perturbations | Attack success rate | Average $L_2$ norm of $p$ | Standard deviation of $p$ |
|---|---|---|---|---|
| Prior k-means classifier | GDU | **1.0000** | 2.1903 | 0.8650 |
| | PSC | 0.9778 | 3.7358 | 1.8549 |
| | OPT | **1.0000** | **0.9633** | **0.3490** |
| Posterior k-means classifier | GDU | **1.0000** | 2.7019 | 1.2333 |
| | PSC | 0.9792 | 4.3079 | 2.4017 |
| | OPT | **1.0000** | **0.8301** | **0.5032** |

The above results show that our proposed evasion attack has high attack success rates with all three algorithms. When using GDU or PSC, the average $L_2$ norm of $p$ is larger than that obtained using OPT. Additionally, the perturbations generated by OPT have the smallest standard deviation of $p$ because GDU and PSC focus on misclassification rather than invisibility, and their performance is relatively worse. Although OPT takes the amount of perturbation into consideration and is more time-consuming, perturbations generated by it have better stealthiness and stability.

We randomly select one sample from each dataset to perform the attack as examples. The visualization of the perturbed images is shown in Figure 3:

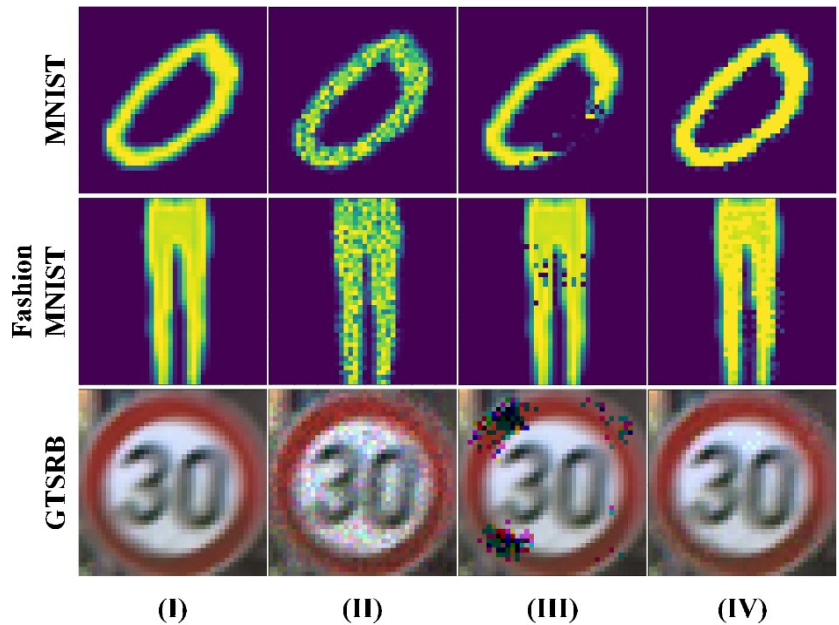

Figure 3 (I) Original images; (II) adversarial samples generated by GDU; (III) adversarial samples generated by

PSC; and (IV) adversarial samples generated by OPT.

Figure 3 shows that for the MNIST and Fashion MNIST samples, the perturbations are confined to the objects and their surrounding areas, meaning that the adversarial images have a similar structure to the original images, which greatly improves the stealthiness of the attack. The experiments on the GTSRB dataset achieve satisfying results as well. When using OPT to generate perturbations based on our method, the perturbations are nearly invisible and the adversarial samples can hardly be distinguished from normal samples; thus, it performs the best among the three algorithms.

The results show that although the SCAE is a type of encoder that encodes the image structure, our attack method, which changes neither the image structure nor the original pose of the existing parts in the image, can still induce a change in the coding result of the SCAE and thus lead to misclassification.

## 6. Conclusion

In this paper, we propose an attack on the state-of-art structure of a capsule network, i.e., the stacked capsule autoencoder. After identifying the object capsule subset related to the original category of the image, an adversarial attack algorithm is used to iteratively compute the perturbation to reduce the presence output by these object capsules. A mask is used when generating the perturbation to improve stealthiness, and the perturbed image will be misclassified by the downstream k-means classifier with a high probability. The experimental results confirm that the SCAE has a security vulnerability in which adversarial samples can be generated without changing the original structure of the image to fool the classifiers. Our work is presented to highlight the threat of this attack and focus attention on SCAE security. In future work, we will study methods to defend against this attack and improve the robustness of the SCAE.